\documentclass[letterpaper]{article} 
\usepackage{aaai25}  
\usepackage{times}  
\usepackage{helvet}  
\usepackage{courier}  
\usepackage[hyphens]{url}  
\usepackage{graphicx} 
\usepackage{natbib}  
\usepackage{caption} 
\usepackage{newfloat}
\usepackage{listings}

\usepackage{booktabs}
\usepackage{subfigure}
\usepackage{amsmath}
\usepackage{amssymb}
\usepackage{tikz}
\usetikzlibrary{fit,positioning}
\usepackage{pgfplots}
\usepackage{enumitem}
\usepackage[ruled,linesnumbered]{algorithm2e}
\usepackage{algpseudocode}
\usepackage{multirow}
\usepackage{makecell}

\DeclareCaptionStyle{ruled}{labelfont=normalfont,labelsep=colon,strut=off} 
\title{Approximated Variational Bayesian Inverse Reinforcement Learning for \\ Large Language Model Alignment}
\author {
    Yuang Cai,
    Yuyu Yuan,
    Jinsheng Shi,
    Qinhong Lin
}
\affiliations {
    Beijing University of Posts and Telecommunications \\
    \{cyang,yuanyuyu,jinsheng,linqinhong\}@bupt.edu.cn
}

\usepackage{bibentry}

\begin{document}

\maketitle

\begin{abstract}
The alignment of large language models (LLMs) is crucial for generating helpful and harmless content.
Existing approaches leverage preference-based human feedback data to learn the reward function and align the LLM with the feedback data.
However, these approaches focus on modeling the reward difference between the chosen and rejected demonstrations, rather than directly modeling the true reward from each demonstration.
Moreover, these approaches assume that the reward is only obtained at the end of the sentence, which overlooks the modeling of intermediate rewards.
These issues lead to insufficient use of training signals in the feedback data, limiting the representation and generalization ability of the reward and potentially resulting in reward hacking.
In this paper, we formulate LLM alignment as a Bayesian Inverse Reinforcement Learning (BIRL) problem and propose a novel training objective, Approximated Variational Alignment (AVA), to perform LLM alignment through Approximated Variational Reward Imitation Learning (AVRIL). 
The BIRL formulation facilitates intermediate reward modeling and direct reward modeling on each single demonstration, which enhances the utilization of training signals in the feedback data.
Experiments show that AVA outperforms existing LLM alignment approaches in reward modeling, RL fine-tuning, and direct optimization.
\end{abstract}

\section{Introduction}

Large language models (LLMs) trained on massive corpus encode a large amount of knowledge and demonstrate powerful linguistic and reasoning capabilities in various domains \cite{openai2022chat,achiam2023gpt}.
However, due to the inevitable harmful and useless information in the training data, LLMs can potentially generate content inconsistent with human values or requirements \cite{holtzman2019curious,zhang2019dialogpt,weidinger2021ethical}.
LLM alignment is a prevalent and effective approach for LLMs to generate harmless and helpful content.
The alignment task typically relies on human feedback data in the form of preferences, where each preference data consists of a chosen sentence and a rejected sentence, labeled by human annotators \cite{zopf2018estimating,tay2020would}.
Reinforcement Learning from Human Feedback (RLHF) and Direct Preference Optimization (DPO) are two common approaches to align LLMs with human feedback data \cite{shen2023large}.
RLHF first performs reward modeling to learn a reward function from the feedback data and then fine-tunes the LLM policy to maximize the expected reward achieved by its generated content using Reinforcement Learning (RL) \cite{ouyang2022training,bai2022training,touvron2023llama,yang2023baichuan}.
DPO formulates the reward modeling objective as a ranking objective based on the LLM policy, which facilitates the joint performance of reward modeling and LLM policy fine-tuning through a unified training objective \cite{yuan2023rrhf,rafailov2024direct,song2024preference}.

The Inverse Reinforcement Learning (IRL) problem generally involves learning a reward model from observed demonstration data produced by a Markov Decision Process (MDP) \cite{ng2000algorithms}.
Conversely, the Natural Language Generation (NLG) process can be viewed as an MDP where the generated sentences are considered demonstration data \cite{ranzato2016sequence}.
Therefore, the alignment task performed by RLHF and DPO can be seen as addressing an IRL problem that infers the implicit reward function hidden in the preference-based human feedback data and learns the LLM policy either separately or jointly.
However, existing RLHF and DPO alignment approaches only model the reward difference between the chosen and rejected demonstrations without explicitly modeling the true reward of every single sentence.
This limitation means that the demonstration data is not fully utilized, which restricts the representation ability of the reward model and can lead to reward hacking \cite{skalse2022defining,gao2023scaling,coste2023reward,zhang2024overcoming}.
Additionally, current approaches generally model the end-to-end sentence-level reward without considering the reward of intermediate states. 
Model generalization may be limited when confronted with data that have similar intermediate state distributions but different complete sentence distributions.
It is more intuitive to model intermediate rewards since humans can not only provide overall feedback on the entire text but also explain which parts of the text influenced their feedback.

In this paper, we propose a novel LLM alignment training objective, \textbf{Approximated Variational Alignment (AVA)}, based on Bayesian Inverse Reinforcement Learning (BIRL) \cite{ramachandran2007bayesian}.
Specifically, we formulate the reward distribution as a posterior distribution conditioned on the demonstration data and perform Approximated Variational Reward Imitation Learning (AVRIL) \cite{chan2021scalable} to jointly approximate the reward distribution (i.e., the reward model) and the demonstration likelihood (i.e., the policy).
Unlike most previous LLM alignment approaches, which only model the reward difference between chosen and rejected demonstrations, AVA directly models the reward of every single demonstration through the AVRIL training objective, thereby making better use of the training signals from feedback data.
Additionally, we do not adhere to the assumption that the reward is only obtained at the end of the sentence.
Instead, we leverage the AVRIL training objective to model the intermediate reward conditioned on the intermediate demonstration data.
To demonstrate the flexibility of our approach, we use the AVA training objective on data in different formats through different pipelines.

Our work makes the following main contributions:
\begin{itemize}
\item We present a novel insight into LLM alignment by formulating the alignment task as a BIRL problem, which enhances the utilization of training signals and improves the representation and generalization ability of the LLM.
\item We demonstrate the flexibility of AVA by employing it for both reward modeling and direct optimization on either preference data or demonstration data.
\item We empirically show that AVA surpasses Bradley-Terry and Preference Transformer in reward modeling and downstream RL fine-tuning, and outperforms DPO and AfD in direct optimization, which indicates a reduction in the reward hacking issue and an improvement in representation and generalization ability.
\end{itemize}

\section{Related Work}

\paragraph{LLM Alignment}
The Bradley-Terry model \cite{bradley1952rank} formulates preference likelihood using the reward model and is widely adopted by RLHF alignment approaches for reward modeling.
After reward modeling, the LLM is fine-tuned to maximize the expected reward achieved by LLM-generated content through downstream RL training \cite{ouyang2022training,bai2022training,touvron2023llama,yang2023baichuan}.
A more concise approach for preference alignment is Direct Preference Optimization (DPO) \cite{rafailov2024direct}, which denotes preference as the relative log-likelihood difference between the chosen sentence and the rejected sentence.
DPO unifies reward modeling and LLM fine-tuning into a single process, facilitating LLM alignment with a simple classification loss.
In addition to aligning LLMs with pairwise human preference data, some recent works also align LLMs with non-pairwise demonstration data.
\citet{sun2024inverse} propose Alignment from Demonstrations (AfD), which leverages high-quality demonstration data to overcome challenges such as noisy labels and privacy concerns in preference datasets.

\paragraph{Intermediate Reward Modeling}
The above alignment approaches only model the end-to-end reward of a complete sentence, without considering the reward of intermediate states.
This lack of intermediate reward modeling stems from the assumption that the reward is only achieved when the sentence is fully generated, regarding the Natural Language Generation (NLG) process as an MDP \cite{ranzato2016sequence}.
To address this issue, we refer to related work on preference modeling in classic RL problems without the aforementioned assumption.
Notably, the Preference Transformer \cite{kim2023preference} uses the attention weights computed by the Transformer architecture \cite{vaswani2017attention} to estimate the weighted non-Markovian reward of each intermediate state of the trajectory.
The reward of a complete trajectory is then the weighted sum of all intermediate rewards.
The preference between the chosen and rejected trajectories is formulated by their rewards and optimized through a contrastive training objective, similar to Bradley-Terry.

\section{Preliminaries}

\subsection{MDP Formulation of NLG}

At time step $t$, the state is the previously generated tokens denoted as $\mathbf{y}_{1:t}=(y_1,y_2,\cdots,y_t)$, the action is the currently generated token $y_{t+1}$.
Note that in auto-regressive decoding, the output tokens are time-shifted.
The action space is the vocabulary $\mathcal{V}$ containing all possible tokens.
In the text generation setting, the state transition is deterministic, so we do not consider the transition probability function.
The reward of taking action $y_{t+1}$ under state $\mathbf{y}_{1:t}$ is denoted as $R(\mathbf{y}_{1:t},y_{t+1})=R(\mathbf{y}_{1:t+1})$, i.e., the reward can be the function of either the current state and the current action or merely the function of the next state due to the deterministic state transition.
It is worth noting that for simplicity of denotation, we do not separately denote the prompt text and the response text but denote them as a whole sentence $\mathbf{y}$.
The separation of prompt and response is trivial during implementation.
The policy can be denoted as $\pi_w(y_{t+1}|\mathbf{y}_{1:t})$, which is also the distribution of the language model parameterized by $w$.
For simplicity, we sometimes denote the policy as $\pi_w(\mathbf{y})=\prod_{t=1}^{|\mathbf{y}|-1}\pi_w(y_{t+1}|\mathbf{y}_{1:t})$, where $|\mathbf{y}|$ is the length of sequence $|\mathbf{y}|$.
Note that the accumulated product starts from $\pi_w(y_2|\mathbf{y}_{1:1})$ instead of $\pi_w(y_1)$ since we assume that all sequences start with a special token denoting the start of the sequence.

\subsection{Bayesian Inverse Reinforcement Learning}

Inverse Reinforcement Learning (IRL) is the problem of extracting a reward function of a Markov Decision Process (MDP) given observed optimal behavior \cite{ng2000algorithms}.
Bayesian Inverse Reinforcement Learning (BIRL) regards the reward function $R$ as the hidden variable affecting and motivating the behavioral data $\mathcal{T}$.
The objective of BIRL is to learn the posterior distribution $p(R|\mathcal{T})$.
Approximate Variational Reward Imitation Learning (AVRIL) \cite{chan2021scalable} adopts variational inference to approximate the posterior distribution.
Specifically, AVRIL employs a parameterized distribution $q_\phi$ and minimizes the Kullback-Leibler (KL) divergence between $q_\phi$ and the posterior distribution $p(R|\mathcal{T})$, as shown in Eq. \ref{eq:avril_min_kl}.
This KL divergence is hard to compute since the posterior distribution is intractable.
A common solution is to maximize the Evidence Lower Bound (ELBO), as shown in Eq. \ref{eq:avril_elbo}, where the second term is to minimize the KL divergence between $q_\phi$ and the tractable prior distribution.

\begin{equation}
\label{eq:avril_min_kl}
\min_\phi D_\text{KL}[q_\phi(R)||p(R|\mathcal{T})]
\end{equation}

\begin{equation}
\label{eq:avril_elbo}
\max_\phi\mathbb{E}_{R\sim q_\phi(\cdot)}[\log p(\mathcal{T}|R)]-D_\text{KL}[q_\phi(R)||p(R)]
\end{equation}

The first term of Eq. \ref{eq:avril_elbo} is to maximize the log-likelihood of the observed optimal behaviors given any reward sampled from $q_\phi$.
AVRIL denotes the action distribution as a Boltzmann policy, as shown in Eq. \ref{eq:avril_elbo_boltz}, where $Q^{\pi_\mathcal{T}}_R$ is the state-action value function following policy $\pi_\mathcal{T}$ under reward function $R$.
Intuitively, we can approximate the state-action value using a Deep Q Network (DQN) \cite{mnih2013playing} $Q_\theta$ parameterized by $\theta$.
An important problem is that in the RL setting, the reward function is fixed when optimizing $Q_\theta$.
However, in the AVRIL setting, the reward function is also being optimized during the optimization of $Q_\theta$.
The reward function and the state-action value function should satisfy $R(s, a)=\mathbb{E}_{s'\sim P(\cdot|s, a),a'\sim\pi(\cdot|s')}[Q^\pi_R(s, a)-\gamma Q^\pi_R(s', a')],~\forall~s\in\mathcal{S},a\in\mathcal{A}$, i.e., the reward should equal the expectation of the TD error.
By adding a penalty term forcing the TD error to follow the reward distribution, the final objective to be maximized is shown in Eq. \ref{eq:avril_final_obj}, where $q_\phi(R|s, a)$ denotes the distribution of reward values given the state $s$ and the action $a$.
In this way, the behavior is indirectly conditioned on the reward, which is consistent with the likelihood $p(\mathcal{T}|R)$ in the ELBO (Eq. \ref{eq:avril_elbo}).
Here, $B(a|s;Q_\theta)$ is the Boltzmann policy upon the state-action value function $Q_\theta$ parameterized by $\theta$.
The third term in the square brackets is to restrict the TD error to satisfy the constraint $R(s, a)=\mathbb{E}_{s',a'}[Q^\pi_R(s, a)-\gamma Q^\pi_R(s', a')],~\forall~s\in\mathcal{S},a\in\mathcal{A}$.
$q_\phi(R|s, a)$ denotes the distribution of reward values given the state $s$ and the action $a$.
We refer to the training objective in Eq. \ref{eq:avril_final_obj} as the approximated variational training objective of the ELBO.

\begin{equation}
\label{eq:avril_elbo_boltz}
B(a|s;Q^{\pi_\mathcal{T}}_R) = \frac{\exp(\beta Q^{\pi_\mathcal{T}}_R(s,a))}{\sum_{a'\in\mathcal{A}}\exp(\beta Q^{\pi_\mathcal{T}}_R(s,a'))}
\end{equation}

\begin{equation}
\label{eq:avril_final_obj}
\begin{aligned}
\max_{\phi,\theta}\sum_{(s,a,s',a')\in\mathcal{T}}\left[
\begin{aligned}
\log B(a|s;Q_\theta)-D_\text{KL}\left[q_\phi(\cdot|s,a)||p(\cdot)\right] \\
+\lambda \log q_\phi\big(Q_\theta(s,a)-\gamma Q_\theta(s',a')|s,a\big)
\end{aligned}
\right]
\end{aligned}
\end{equation}

\section{Approximated Variational Alignment}

In this section, we formulate the LLM alignment tasks as the BIRL problems and perform alignment with the Approximated Variational Alignment (AVA) training objectives.
The AVA training objectives involve AVA from Demonstration (AVA-d) and AVA from Preference (AVA-p), both of which are BIRL training objectives based on the Approximated Variational Reward Imitation Learning (AVRIL) training objective \cite{chan2021scalable}.
AVA-d is the implementation of the AVRIL training objective under the NLG setting, which learns on non-pairwise demonstration datasets.
AVA-p is a contrastive variant of AVA-d, which learns on pairwise preference datasets.

\subsection{Alignment from Demonstration}

We first consider the problem of aligning an LLM policy with the demonstration data $\mathcal{D}$, where each sentence $\mathbf{y}\in\mathcal{D}$ is the ground-truth sentence.
The alignment objective is to encourage the LLM policy to generate sentences like the demonstration data.
Instead of building a direct training objective (e.g., supervised fine-tuning) to optimize the LLM policy, we focus on performing BIRL to learn a reward function from the demonstration data $\mathcal{D}$, i.e., to learn the posterior $p(R|\mathcal{D})$ with a parameterized distribution $q_\phi(R)$.

As illustrated in the preliminaries, the optimization of $q_\phi$ can be achieved by maximizing the AVRIL training objective (Eq. \ref{eq:avril_final_obj}), where each element $(s,a,s',a')\in\mathcal{T}$ is a state-action quadruplet consisting of the current state $s$, the current action $a$, the next state $s'$ and the next action $a'$.
As for the Natural Language Generation (NLG) setting, at each time step $t$, the current state is the current sub-sentence $\mathbf{y}_{1:t}$, the current action is the token-to-be-generated $y_{t+1}$, the next state is $\mathbf{y}_{1:t+1}$, the concatenation of $\mathbf{y}_{1:t}$ and $y_{t+1}$, and the next action is $y_{t+2}$.
By substituting the state-action quadruplet in Eq. \ref{eq:avril_final_obj} with the new quadruplet $(\mathbf{y}_{1:t},y_{t+1},\mathbf{y}_{1:t+1},y_{t+2})$ and rewrite the summation in timestep-wise form, we can obtain the AVRIL training objective applicable to the NLG setting, as shown in Eq. \ref{eq:avril_obj_corpus}.
We refer to this training objective as Approximated Variational Alignment from Demonstration (AVA-d), which is a variant of the AVRIL training objective in the natural language generation setting.

\begin{equation}
\label{eq:avril_obj_corpus}
\mathcal{F}_d(\mathcal{D})=\sum_{\mathbf{y}\in\mathcal{D}}\sum_{t=1}^{|\mathbf{y}|-2}\left[
\begin{aligned}
\log B(y_{t+1}|\mathbf{y}_{1:t};Q_\theta)
-d_t(\phi) \\
+\lambda\log q_\phi\big(\delta_t(\theta)|\mathbf{y}_{1:t+1}\big)
\end{aligned}
\right]
\end{equation}

\begin{equation}
d_t(\phi) = D_\text{KL}\left[q_\phi(\cdot|\mathbf{y}_{1:t+1})||p(\cdot)\right]
\end{equation}

\begin{equation}
\label{eq:avril_td_error}
\delta_t(\theta) =Q_\theta(\mathbf{y}_{1:t},y_{t+1})-\gamma Q_\theta(\mathbf{y}_{1:t+1},y_{t+2})
\end{equation}

Here, $q_\phi(R|\mathbf{y}_{1:t+1})$ is the reward distribution of the sub-sequence $\mathbf{y}_{1:t+1}$.
The Boltzmann policy $B(y_{t+1}|\mathbf{y}_{1:t}; Q_\theta)$ built upon the Q-value model $Q_\theta$ acts as the LLM policy for text generation.
By maximizing $\mathcal{F}_d$, the Q-value model (i.e., the LLM policy) $Q_\theta$ as well as the reward distribution $q_\phi$ will be jointly optimized to be aligned with the demonstration dataset $\mathcal{D}$.

Similar to the original AVRIL objective, the AVA-d objective consists of three sub-objectives: the log-likelihood maximization, the KL divergence minimization, and the TD-error constraint.
The first objective trains the LLM policy to maximize the likelihood of the demonstration data, which is identical to supervised fine-tuning.
The second objective is to ensure the reward distribution satisfies the prior distribution assumption.
The third objective, TD-error constraint, distinguishes AVA-d from conventional supervised fine-tuning.
With the constraint, the update of the Q-value model will not only increase the Q-value of the ground-truth token in demonstration data but also make the TD error of the Q-values \ref{eq:avril_td_error} close to the reward obtained after generating the current token, which ensures the consistency between the reward and the policy.

\subsection{TQR Architecture}

\begin{figure}
\centering
\includegraphics[width=0.6\linewidth]{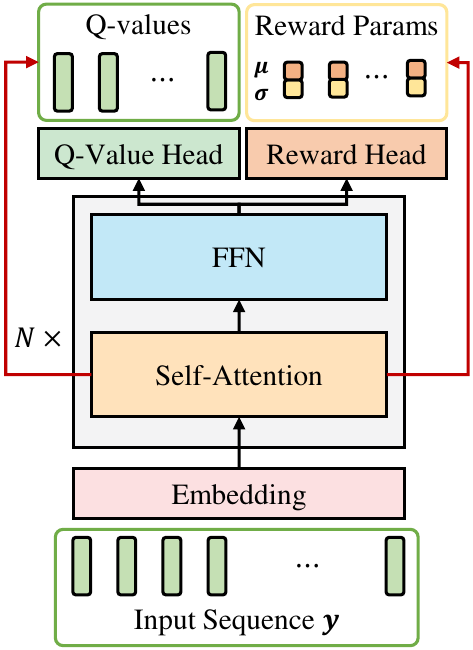}
\caption{Overview of the TQR architecture.}
\label{fig:tqr-arch}
\end{figure}

The original AVRIL adopts the architecture with a reward encoder and a Q-value decoder.
To compute the AVA-d training objective and leverage the pre-trained weights of the backbone transformer model, we add a reward head and a Q-value head at the top of the Transformer decoder, as shown in Fig. \ref{fig:tqr-arch}.
We refer to this architecture as \textbf{T}ransformer with \textbf{Q}-value and \textbf{R}eward Heads (TQR).
The Q-value head takes the hidden states of the last decoder layer as input and outputs the Q-value of each action (i.e., token), as shown in Eq. \ref{eq:q_head_forward}.
The reward is assumed to follow Gaussian distribution, and the reward head takes in the same hidden states and outputs the mean and standard deviation of the reward of each state, as shown In Eq. \ref{eq:r_head_forward}.
Here, $\mathbf{h}_t$ is the hidden state vector of time step $t$, $Q_\theta(\mathbf{y}_{1:t},\cdot)\in\mathbb{R}^{|\mathcal{V}|}$ is a vector whose $i$-th element equals $Q_\theta(\mathbf{y}_{1:t},v^{(i)})$, where $v^{(i)}$ is the $i$-th token in the vocabulary, and $\mu_t, \sigma_t \in \mathbb{R}$ are mean and standard deviation of reward $R(\mathbf{y}_{1:t+1})$ at time step $t$.
Now we can compute the training objective in Eq. \ref{eq:avril_obj_corpus} based on the above outputs of the Q-value head and the reward head.

\begin{align}
\label{eq:q_head_forward}
Q_\theta(\mathbf{y}_{1:t},\cdot) &= \text{QHead}(\mathbf{h}_t;\theta),\forall~t\in\{1,\cdots,|\mathbf{y}|\}\\
\label{eq:r_head_forward}
[{\mu_t};{\sigma_t}]&=\text{RHead}(\mathbf{h}_t;\phi),\forall~t\in\{1,\cdots,|\mathbf{y}|\} \\
\label{eq:r_sample}
R(\mathbf{y}_{1:t+1})&\sim q_\phi(R|\mathbf{y}_{1:t+1})=\mathcal{N}(R;\mu_t,\sigma_t)
\end{align}

Inspired by preference transformer \cite{kim2023preference}, we further compute a reward weight for each time step of reward based on attention weights, as shown in Eq. \ref{eq:reward_weight}, where $\mathbf{q}_i$ is the $i$-th row of the query matrix of the attention mechanism, $\mathbf{k}_{t'}$ is the $t'$-th row of the key matrix.
We then apply reward weights to the outputs of the Q-value head (Eq. \ref{eq:q_head_forward}) and reward head (Eq. \ref{eq:r_head_forward}).
Specifically, we simply multiply the output of the $t$-th position of the heads by the reward weight $w_t$, as shown by the red arrows in Fig. \ref{fig:tqr-arch}.

\begin{equation}
\label{eq:reward_weight}
w_t = \frac{1}{|\mathbf{y}|}\sum_{i=1}^{|\mathbf{y}|}\sum_{t=1}^{|\mathbf{y}|}\text{softmax}\left(\left\{\mathbf{q}_i\cdot\mathbf{k}_{t'}\right\}_{t'=1}^{|\mathbf{y}|}\right)_t
\end{equation}

Besides using a randomly initialized Q-value head, we can also construct a pre-trained Q-value model from the pre-trained LLM policy.
The Boltzmann policy formulates the action probability as the softmax function of Q-values.
Inversely, we can also formulate the Q-value as the log-softmax function of action probabilities, as shown in Eq. \ref{eq:inverse-boltzmann-q}, where $\alpha$ is the temperature hyperparameter, $\pi_w$ is the LLM policy parameterized by $w$.
Note that the log-softmax operation is a non-strict inversion of the softmax operation, which means we can tune $\alpha$ to find the best way to map token-level probabilities to token-level Q-values.

\begin{equation}
\label{eq:inverse-boltzmann-q}
Q_w(\mathbf{y}_{1:t},y_{t+1})=\log\frac{\exp(\alpha\pi_w(y_{t+1}|\mathbf{y}_{1:t}))}{\sum_{y'\in\mathcal{V}}\exp(\alpha \pi_w(y'|\mathbf{y}_{1:t}))}
\end{equation}

By substituting with the above Q-value model, the AVA-d training objective can be denoted as Eq. \ref{eq:avril-obj}, and the TD error can be denoted as Eq. \ref{eq:td-error-q}.
This denotation facilitates us to initialize the Q-value model from a pre-trained LLM policy and adopt the AVA-d objective to fine-tune the LLM policy.

\begin{equation}
\label{eq:avril-obj}
\mathcal{F}_d(\mathcal{D})=\sum_{\mathbf{y}\in\mathcal{D}}\sum_{t=1}^{|\mathbf{y}|-2}\left[
\begin{aligned}
\beta\log\text{softmax}(\alpha\pi_w(y_{t+1}|\mathbf{y}_{1:t})) \\
-d_t(\phi) 
+\lambda\log q_\phi(\delta_t(w)|\mathbf{y}_{1:t+1})
\end{aligned}
\right]
\end{equation}

\begin{equation}
\label{eq:td-error-q}
\delta_t(w)=\log\frac{\text{softmax}(\alpha\pi_w(y_{t+1}|\mathbf{y}_{1:t}))}{\text{softmax}(\alpha\pi_w(y_{t+2}|\mathbf{y}_{1:t+1}))^\gamma}
\end{equation}

\subsection{Alignment from Preference}

We then consider the problem of aligning an LLM policy $\pi_w$ with preference data $\mathcal{P}$, where each data item $(\mathbf{y}^+,\mathbf{y}^-)\in\mathcal{P}$ consists of the chosen sentence $\mathbf{y}^+$ and the rejected sentence $\mathbf{y}^-$.
We denote the set of all chosen sentences as $\mathcal{P}^+=\{\mathbf{y}^+|(\mathbf{y}^+,\mathbf{y}^-)\in\mathcal{P}\}$ and the set of all rejected sentences as $\mathcal{P}^-=\{\mathbf{y}^-|(\mathbf{y}^+,\mathbf{y}^-)\in\mathcal{P}\}$.
The alignment objective is to encourage the LLM policy to generate sentences like the chosen demonstrations $\mathcal{P}^+$ while discouraging the LLM policy from generating sentences like the rejected demonstrations $\mathcal{P}^-$.

Similar to the derivation of the AVA-d training objective, we first focus on performing BIRL to learn a reward function from the preference data $\mathcal{P}$.
We need to consider not only the chosen sentences as positive demonstrations but also the rejected sentences as negative demonstrations.
We consider two posterior distributions, which are the reward conditioned on the chosen demonstrations $p(R|\mathcal{P}^+)$ and the reward conditioned on demonstrations that differ from rejected demonstrations $p(R|\overline{\mathcal{P}^-})$.
Here, $\overline{\mathcal{P}^-}$ denotes demonstrations that differ from $\mathcal{P}^-$.
Therefore, we define the training objective as Eq. \ref{eq:contrast-birl}, where the first term drives the reward distribution $q_\phi$ close to rewards that motivate the positive behaviors $\mathcal{P}^+$, while the second term drives $q_\phi$ close to rewards that motivate behaviors that differ from the negative demonstrations.
We refer to the training objective as Contrastive Bayesian Inverse Reinforcement Learning (CBIRL).

\begin{equation}
\label{eq:contrast-birl}
\min_\phi D_\text{KL}[q_\phi(R)||p(R|\mathcal{P}^+)] + D_\text{KL}[q_\phi(R)||p(R|\overline{\mathcal{P}^-})]
\end{equation}

Unsurprisingly, the minimization of these two KL divergences is infeasible.
We derive the equivalent ELBO objective, as shown in Eq. \ref{eq:elbo-cbirl}.
The derivation is shown in the Technical Appendix.

\begin{equation}
\label{eq:elbo-cbirl}
\max_\phi\left[
\begin{aligned}
\mathbb{E}_{R\sim q_\phi(\cdot)}\left[\log p(\mathcal{P}^+|R)+\log[1-p(\mathcal{P}^-|R)]\right]\\
-D_\text{KL}[q_\phi(R)||p(R)]
\end{aligned}
\right]
\end{equation}

Towards implementation, we need to further derive the ELBO objective as an approximated variational objective.
Note that the main difference between the ELBO of CBIRL and the ELBO of conventional BIRL is the second optimization term in Eq. \ref{eq:elbo-cbirl}, which minimizes the log-likelihood of the negative demonstrations $\mathcal{P}^-$.
Therefore, the approximated variational objective also contains the minimization of the negative demonstrations, as shown in Eq. \ref{eq:cavril-obj}.
We refer to this training objective as the Approximated Variational Alignment from Preference (AVA-p).
By maximizing $\mathcal{F}_p(\mathcal{P})$, on the one hand, the LLM policy $\pi_w$ will be encouraged to generate sentences like $\mathcal{P}^+$ and discouraged to generated sentences like $\mathcal{P}^-$; on the other hand, the policy and the reward will stay consistent under the TD-error constraint.

\begin{equation}
\label{eq:cavril-obj}
\mathcal{F}_p(\mathcal{P})=\sum_{\mathbf{y}^{+/-}\in\mathcal{P}}\sum_{t}\left[
\begin{aligned}
\beta\log\text{softmax}(\alpha\pi_w(y_{t+1}^+|\mathbf{y}_{1:t}^+)) \\
-\beta\log\text{softmax}(\alpha\pi_w(y_{t+1}^-|\mathbf{y}_{1:t}^-)) \\
-d_t(\phi)+\lambda\log q_\phi\big(\delta_t(\theta)|\mathbf{y}_{1:t+1}\big)
\end{aligned}
\right]
\end{equation}

To ensure the reward difference between the chosen and rejected demonstrations, we adopt a more intuitive auxiliary training objective, the Contrastive Expected Return (CER) training objective,  as shown in Eq. \ref{eq:cer}, which encourages the reward of the positive demonstrations to be higher than the reward of the negative demonstrations.
Note that we only consider the reward of the last timestep in the CER objective.
Although we model the intermediate rewards, we still assume that the reward of the last timestep is decisive for the overall expected return, since empirical practice and research \cite{geva2023dissecting,hanna2024does} show that the last position of the Transformer gathers most of the knowledge.

\begin{equation}
\label{eq:cer}
\begin{aligned}
\mathcal{F}_c(\mathcal{P})&=\sum_{\mathbf{y}^{+/-}\in\mathcal{P}}\sigma\left[\mathbb{E}_{q_\phi(R|\mathbf{y}^+)}[R]-\mathbb{E}_{q_\phi(R|\mathbf{y}^-)}[R]\right]
\end{aligned}
\end{equation}

\subsection{AVA Pipelines}

The AVA training objectives facilitate the joint optimization of the reward function and the policy.
Therefore, AVA can be leveraged for both reward modeling and direct optimization, which are two common pipelines in LLM alignment.
Both pipelines have their advantages and disadvantages.
The reward modeling pipeline can produce a lightweight and reusable reward function for downstream RL fine-tuning while it suffers from the high RL training cost.
The direct optimization pipeline is more efficient than reward modeling with RL during training but cannot produce a lightweight reward function for other uses and may suffer from overfitting.

\subsubsection{AVA for Reward Modeling}

\begin{algorithm}
\caption{AVA for reward modeling.}
\label{algo:rm-rlft}
\KwData{Dataset $\mathcal{D}$, initial implicit policy $\pi_{\psi^{(1)}}$, initial reward distribution $q_{\phi^{(1)}}$, training epochs $T$}
\KwResult{The trained reward distribution $q_{\phi^{(T)}}$}
\For{$i\in\{1,\cdots,T\}$}{
  \eIf{$\mathcal{D}$ is demonstration dataset}{
    $\phi^{(i+1)}\leftarrow\phi^{(i)}+\nabla_{\phi^{(i)}}\mathcal{F}_d(\mathcal{D})$\;
    $\psi^{(i+1)}\leftarrow\psi^{(i)}+\nabla_{\psi^{(i)}}\mathcal{F}_d(\mathcal{D})$\;
  }{
    $\phi^{(i+1)}\leftarrow\phi^{(i)}+\nabla_{\phi^{(i)}}\mathcal{F}_p(\mathcal{D})+\nabla_{\phi^{(i)}}\mathcal{F}_c(\mathcal{D})$\;
    $\psi^{(i+1)}\leftarrow\psi^{(i)}+\nabla_{\psi^{(i)}}\mathcal{F}_p(\mathcal{D})+\nabla_{\psi^{(i)}}\mathcal{F}_c(\mathcal{D})$\;
  }
}
\Return{$q_{\phi^{(T)}}$}
\end{algorithm}

The pipeline of AVA for reward modeling is shown in Alg. \ref{algo:rm-rlft}.
In the TQR architecture, the reward function shares the same backbone model with the policy.
Our purpose of reward modeling is to obtain an accurate and lightweight reward model.
Therefore, we initialize the TQR architecture with a lightweight backbone model.
In other words, we initialize the policy with a lightweight pre-trained language model $\pi_{\psi^{(1)}}$ instead of a large language model.
Meanwhile, the reward distribution is also initialized and denoted by $q_{\phi^{(1)}}$
After the initialization, we leverage either AVA-d or AVA-p training objectives to optimize the reward distribution according to the type of the dataset $\mathcal{D}$.
Note that the AVA training objectives require us to jointly train the reward function with the policy, although finally we only need the reward function.
After reward modeling, we can leverage RL algorithms to fine-tune the LLM policy $\pi_w$ to maximize the expected reward produced by the trained reward distribution $q_\phi$, as shown in Eq. \ref{eq:rlft-sotch-reward-obj-approx}.

\begin{equation}
\label{eq:rlft-sotch-reward-obj-approx}
J(w)=\mathbb{E}_{\mathbf{y}\sim\pi_w(\cdot)}\left[\sum_{t=1}^{|\mathbf{y}|-1}\mathbb{E}_{R\sim q_\phi(\cdot|\mathbf{y}_{1:t+1})}[R]\right]
\end{equation}

\subsubsection{AVA for Direct Optimization}

The pipeline of AVA for direct optimization is shown in Alg. \ref{algo:dio}.
Here, we directly initialize the policy with the pre-trained LLM $\pi_{w^{(1)}}$ and leverage the AVA training objectives to jointly optimize the policy and the reward distribution $q_{\phi^{(1)}}$.
After training, the LLM policy and the reward distribution are both aligned with the demonstration or preference dataset $\mathcal{D}$.

\begin{algorithm}
\caption{AVA for Direct Optimization.}
\label{algo:dio}
\KwData{Dataset $\mathcal{D}$, initial LLM policy $\pi_{w^{(1)}}$, initial reward distribution $q_{\phi^{(1)}}$, training epochs $T$}
\KwResult{The finally trained LLM policy $\pi_{w^{(T)}}$}
\For{$i\in\{1,\cdots,T\}$}{
  \eIf{$\mathcal{D}$ is demonstration dataset}{
    $\phi^{(i+1)}\leftarrow\phi^{(i)}+\nabla_{\phi^{(i)}}\mathcal{F}_d(\mathcal{D})$\;
    $w^{(i+1)}\leftarrow w^{(i)}+\nabla_{w^{(i)}}\mathcal{F}_d(\mathcal{D})$\;
  }{
    $\phi^{(i+1)}\leftarrow\phi^{(i)}+\nabla_{\phi^{(i)}}\mathcal{F}_p(\mathcal{D})+\nabla_{\phi^{(i)}}\mathcal{F}_c(\mathcal{D})$\;
    $w^{(i+1)}\leftarrow w^{(i)}+\nabla_{w^{(i)}}\mathcal{F}_p(\mathcal{D})+\nabla_{w^{(i)}}\mathcal{F}_c(\mathcal{D})$\;
  }
}
\Return{$\pi_{w^{(T)}}$}\;
\end{algorithm}

\section{Experiment}

\subsection{Experiment Setup}

\paragraph{Datasets}

For preference datasets, we consider Anthropic-Harmless, Anthropic-Helpful, and OpenAI-Summary and perform reward modeling, RL fine-tuning, and direct optimization on these datasets.
For demonstration datasets, we consider Alpaca-GPT-4 and Math-GPT-4o and only perform direct optimization on these datasets.

\paragraph{Metrics}

For reward modeling, we evaluate the accuracy at which the reward of the chosen sentence is greater than that of the rejected sentence, as well as the win rates of the Best-of-N sampling \cite{stiennon2020learning,nakano2021webgpt} results.
For RL fine-tuning, we evaluate the win rates of the LLMs fine-tuned with different reward models (i.e., AVA-p/d and baselines).
For direct optimization, we evaluate the win rates of LLMs fine-tuned with AVA-p/d against LLMs fine-tuned with baseline approaches.

\paragraph{Pre-trained Models}

For reward modeling, we initialize the implicit policy with GPT-2 (117M) and BART-base (140M) to see the reward modeling performance with different initializations.
For RL fine-tuning and direct optimization, we initialize the LLM policy with Llama-2-7b-chat-hf.
The reward models adopted in RL fine-tuning only involve those initialized with GPT-2.

\paragraph{Baselines}

For reward modeling, we adopt Bradley-Terry \cite{bradley1952rank} and Preference Transformer (Pref-Trans) \cite{kim2023preference} as baselines.
For direct optimization from preference, we adopt DPO \cite{rafailov2024direct} as the baseline.
For direct optimization from demonstration, we adopt AfD \cite{sun2024inverse} as the baseline.
Since AfD constructs preference data from demonstration data and relies on preference-based training objectives, we combine AfD with different preference-based training objectives.
Specifically, for reward modeling, we construct AfD w/ Bradley-Terry, Afd w/ Pref-Trans, and AfD w/ AVA-p.
For direct optimization, we construct AfD w/ DPO.
For win rate evaluations of aligned LLMs, we also adopt supervised fine-tuning (SFT) as the baseline.

\paragraph{Ablation Variants}

We construct the following variants of the AVA-p and AVA-d training objectives for ablation studies:
\begin{itemize}
\item \textbf{AVA-p/d w/o rwt}: AVA-p/d without reward weighting, which removes the computation of reward weights and the weighted rewards from the TQR architecture.
\item \textbf{AVA-p w/o neg}: AVA-p without the negative demonstration, which removes the minimization of the likelihood of the negative demonstrations. Note that the objective does not completely degenerate into the AVA-d training objective since we still keep the CER auxiliary objective.
\item \textbf{AVA-p w/o irl}: AVA-p without inverse reinforcement learning, which removes the TD-error constraint and the reward prior assumption and only keeps the likelihood optimization, which can be regarded as contrastive supervised fine-tuning.
\item \textbf{AVA-p w/o cer}: AVA-p without CER auxiliary objective.
\item \textbf{AVA-p/d w/o ptq}: AVA-p/d without pre-trained Q-value head, which does not reuse the LM head of the pre-trained policy as the Q-value head but initializes the Q-value head from scratch.
\end{itemize}

For detailed experiment setup, please refer to our code and the Experiment Details section of the Technical Appendix.

\subsection{Reward Modeling}

\begin{table}
\setlength\tabcolsep{3pt}
\centering
\begin{tabular}{@{}lrrrrrr@{}}
\toprule
                    & \multicolumn{2}{c}{\textbf{Harmless}}                   & \multicolumn{2}{c}{\textbf{Helpful}}                    & \multicolumn{2}{c}{\textbf{Summary}}                    \\
                    & gpt2                       & bart                       & gpt2                       & bart                       & gpt2                       & bart                       \\ \midrule
\textit{Baselines}  &                            &                            &                            &                            &                            &                            \\
Bradley-Terry       & 70.02                      & 68.96                      & 69.39                      & 67.56                      & 59.27                      & 59.27                      \\
Pref-Trans          & 70.26                      & 71.32                      & 71.37                      & 72.37                      & 59.31                      & 56.91                      \\ \midrule
\textit{Ours}       &                            &                            &                            &                            &                            &                            \\
AVA-p               & \underline{70.27}          & \underline{\textbf{72.30}} & \underline{\textbf{72.37}} & \underline{\textbf{74.84}} & \underline{61.79}          & \underline{\textbf{64.31}} \\
AVA-p w/o rwt       & 70.06                      & 70.73                      & 69.81                      & 69.32                      & \underline{60.55}          & 58.89                      \\
AVA-p w/o neg       & \underline{\textbf{70.54}} & 70.36                      & 69.75                      & 69.15                      & \underline{\textbf{62.06}} & 58.65                      \\
AVA-p w/o irl       & 69.48                      & 67.46                      & 68.87                      & 65.38                      & 58.96                      & 58.46                      \\
AVA-p w/o cer       & 70.06                      & 70.73                      & 69.81                      & 69.32                      & \underline{60.55}          & 58.58                      \\
AVA-p w/o ptq       & 68.67                      & 68.69                      & 68.51                      & 67.60                      & \underline{61.25}          & 57.76                      \\
AVA-d               & \underline{\textbf{70.54}} & 70.36                      & 69.75                      & 69.15                      & \underline{\textbf{62.06}} & 59.00                      \\ \bottomrule
\end{tabular}
\caption{Reward accuracy of AVA and baseline training objectives.}
\label{tab:reward-modeling-accuracy}
\end{table}

Table \ref{tab:reward-modeling-accuracy} reports the reward accuracy of baseline and AVA training objectives.
The results show that AVA-p surpasses Bradley-Terry and Pref-Trans in reward accuracy on all reported reward modeling tasks with different initial models and datasets.
The ablation results further reveal that AVA-p achieves the highest reward accuracy on the greatest number of tasks compared to ablated training objectives, which suggests that removing any module from AVA-p diminishes the reward accuracy on most tasks.
Furthermore, we consider the chosen half of the preference data as demonstration data and train the reward model on it using the AVA-d training objective.
Surprisingly, AVA-d achieves the best performance on 2 out of 6 tasks, despite learning solely from the chosen demonstrations.

To further evaluate reward modeling performance, we employ Best-of-N (BoN) sampling.
We evaluate the win rates of BoN w/ AVA-p against BoN w/ Bradley-Terry and BoN w/ Pref-Trans, where “BoN w/ xxx” means that the reward model used for BoN is trained with the “xxx” training objective.
Additionally, we evaluate the win rate of BoN w/ AVA-p against the stochastic sampling results without BoN.
Table \ref{tab:bon} reports the win rates of the reward model trained with AVA-p against reward models trained with baseline objectives in BoN sampling.
The results further demonstrate that AVA-p surpasses Bradley-Terry and Pref-Trans in reward modeling.

\begin{table}
\setlength\tabcolsep{3pt}
\centering
\begin{tabular}{@{}clrrr@{}}
\toprule
\multicolumn{1}{l}{\textbf{Task}} & \textbf{Opponent}    & \textbf{Win↑} & \textbf{Tie} & \textbf{Lose↓} \\ \midrule
\multirow{3}{*}{Harmless}         & Stochastic           & 43.0          & 17.4         & 39.6           \\
                                  & BoN w/ Bradley-Terry & 28.8          & 42.6         & 28.6           \\
                                  & BoN w/ Pref-Trans    & 35.6          & 31.9         & 32.5           \\ \midrule
\multirow{3}{*}{Helpful}          & Stochastic           & 26.1          & 50.4         & 23.5           \\
                                  & BoN w/ Bradley-Terry & 13.2          & 76.1         & 10.7           \\
                                  & BoN w/ Pref-Trans    & 19.3          & 62.3         & 18.4           \\ \midrule
\multirow{3}{*}{Summary}          & Stochastic           & 60.2          & 0.8          & 39.0           \\
                                  & BoN w/ Bradley-Terry & 34.6          & 34.5         & 30.9           \\
                                  & BoN w/ Pref-Trans    & 43.6          & 25.8         & 30.6           \\ \bottomrule
\end{tabular}
\caption{Win rates of BoN with AVA-p reward model.}
\label{tab:bon}
\end{table}

\subsection{RL Fine-tuning}

\begin{table}
\setlength\tabcolsep{3pt}
\centering
\begin{tabular}{@{}clrrr@{}}
\toprule
\multicolumn{1}{l}{\textbf{Task}} & \textbf{Opponent}    & \textbf{Win↑} & \textbf{Tie} & \textbf{Lose↓} \\ \midrule
\multirow{3}{*}{Harmless}         & SFT                  & 42.5          & 23.4         & 34.1           \\
                                  & PPO w/ Bradley-Terry & 9.2           & 81.7         & 9.1            \\
                                  & PPO w/ Pref-Trans    & 9.0           & 83.2         & 7.8            \\ \midrule
\multirow{3}{*}{Helpful}          & SFT                  & 23.3          & 58.8         & 18.0           \\
                                  & PPO w/ Bradley-Terry & 1.8           & 97.2         & 1.0            \\
                                  & PPO w/ Pref-Trans    & 2.6           & 95.8         & 1.6            \\ \midrule
\multirow{3}{*}{Summary}          & SFT                  & 73.8          & 1.4          & 24.7           \\
                                  & PPO w/ Bradley-Terry & 18.5          & 66.3         & 15.2           \\
                                  & PPO w/ Pref-Trans    & 33.9          & 34.6         & 31.5           \\ \bottomrule
\end{tabular}
\caption{Win rates of PPO with AVA-p reward model.}
\label{tab:rm-rlft-wtl}
\end{table}

We adopt the PPO algorithm \cite{schulman2017proximal} to fine-tune LLMs to maximize the reward produced by different reward models.
We evaluate the win rates of PPO w/ AVA-p against PPO w/ Bradley-Terry and PPO w/ Pref-Trans, where “PPO w/ xxx” means that the reward model used for PPO fine-tuning is trained with the “xxx” training objective.
We also evaluate the win rate of PPO w/ AVA-p against supervised fine-tuning (SFT), where the LLM is fine-tuned on the chosen half of the preference data with supervised learning.
The results in Table \ref{tab:rm-rlft-wtl} show that AVA-p outperforms the baseline reward modeling objectives on all reported tasks in downstream RL fine-tuning of the LLM.

\subsection{Direct Optimization}

\begin{table}[]
\centering
\begin{tabular}{@{}clrrr@{}}
\toprule
\textbf{Task}             & \textbf{Opponent} & \textbf{Win↑} & \textbf{Tie} & \textbf{Lose↓} \\ \midrule
\multirow{2}{*}{Harmless} & SFT               & 37.1          & 28.9         & 34.0           \\
                          & DPO               & 13.7          & 73.8         & 12.5           \\ \midrule
\multirow{2}{*}{Helpful}  & SFT               & 22.5          & 59.6         & 17.9           \\
                          & DPO               & 14.4          & 72.4         & 13.2           \\ \midrule
\multirow{2}{*}{Summary}  & SFT               & 59.0          &  7.3         & 33.7           \\
                          & DPO               & 44.9          & 11.0         & 44.1           \\ \bottomrule
\end{tabular}
\caption{Win rates of direct optimization with AVA-p.}
\label{tab:do-preference}
\end{table}

\paragraph{From Preference}

We adopt AVA-p and DPO \cite{rafailov2024direct} to directly optimize the LLM from preference data and evaluate the win rates of AVA-p against DPO and SFT.
The results in Table \ref{tab:do-preference} show that AVA-p outperforms DPO in direct optimization from preference data.

\paragraph{From Demonstration}

We adopt AVA-d and AfD \cite{sun2024inverse} to directly optimize the LLM from demonstration data.
We evaluate the win rates of AVA-d against SFT, DPO w/ AfD, and AVA-p w/ AfD, where “xxx w/ AfD” means applying the “xxx” training objective on AfD-format data.
The results in Table \ref{tab:direct-demon-wtl} show that AVA-d outperforms the AfD approaches in direct optimization from demonstration data.
Moreover, AVA-d is more training-efficient since AfD requires supervised fine-tuning and sampling from LLM policies.

\begin{table}
\centering
\begin{tabular}{@{}clrrr@{}}
\toprule
\multicolumn{1}{l}{\textbf{Task}}   & \textbf{Opponent} & \textbf{Win↑} & \textbf{Tie} & \textbf{Lose↓} \\ \midrule
\multirow{3}{*}{Alpaca}             & SFT               & 58.1          &  7.2         & 34.7           \\
                                    & DPO w/ AfD        & 57.2          &  6.9         & 35.9           \\
                                    & AVA-p w/ AfD      & 56.5          &  7.1         & 36.4           \\ \midrule
\multirow{3}{*}{Math}               & SFT               & 47.0          &  9.7         & 43.3           \\
                                    & DPO w/ AfD        & 44.3          & 11.4         & 44.3           \\
                                    & AVA-p w/ AfD      & 45.4          & 11.4         & 43.1           \\ \bottomrule
\end{tabular}
\caption{Win rates of direct optimization with AVA-d.}
\label{tab:direct-demon-wtl}
\end{table}

\section{Conclusion}

We present AVA, a flexible novel LLM alignment objective with enhanced capabilities.
The flexibility of AVA is evident in two aspects.
Firstly, AVA can utilize either preference data or demonstration data for alignment purposes.
Secondly, AVA can be integrated into the reward modeling and RL fine-tuning pipeline or used to directly optimize the LLM.
The representation and generalization capabilities of AVA are also evident in two aspects.
Theoretically, AVA formulates reward modeling as a BIRL problem, facilitating both intermediate reward modeling and direct reward modeling on demonstration.
Experimentally, AVA achieves superior reward accuracy in reward modeling tasks and higher win rates in RL fine-tuning and direct optimization of LLMs, which demonstrates the alleviation of the reward hacking issue and improved alignment performance.

\bibliography{aaai25}

\section{Reproducibility Checklist}

This paper:

\begin{itemize}
\item Includes a conceptual outline and/or pseudocode description of AI methods introduced (yes)
\item Clearly delineates statements that are opinions, hypothesis, and speculation from objective facts and results (yes)
\item Provides well marked pedagogical references for less-familiare readers to gain background necessary to replicate the paper (yes)
\end{itemize}

Does this paper make theoretical contributions? (yes)

If yes, please complete the list below.

\begin{itemize}
\item All assumptions and restrictions are stated clearly and formally. (yes)
\item All novel claims are stated formally (e.g., in theorem statements). (yes)
\item Proofs of all novel claims are included. (yes)
\item Proof sketches or intuitions are given for complex and/or novel results. (yes)
\item Appropriate citations to theoretical tools used are given. (yes)
\item All theoretical claims are demonstrated empirically to hold. (yes)
\item All experimental code used to eliminate or disprove claims is included. (yes)
\end{itemize}

Does this paper rely on one or more datasets? (yes)

If yes, please complete the list below.

\begin{itemize}
\item A motivation is given for why the experiments are conducted on the selected datasets (NA)
\item All novel datasets introduced in this paper are included in a data appendix. (NA)
\item All novel datasets introduced in this paper will be made publicly available upon publication of the paper with a license that allows free usage for research purposes. (NA)
\item All datasets drawn from the existing literature (potentially including authors’ own previously published work) are accompanied by appropriate citations. (yes)
\item All datasets drawn from the existing literature (potentially including authors’ own previously published work) are publicly available. (yes)
\item All datasets that are not publicly available are described in detail, with explanation why publicly available alternatives are not scientifically satisficing. (NA)
\end{itemize}

Does this paper include computational experiments? (yes)

If yes, please complete the list below.

\begin{itemize}
\item Any code required for pre-processing data is included in the appendix. (yes).
\item All source code required for conducting and analyzing the experiments is included in a code appendix. (yes)
\item All source code required for conducting and analyzing the experiments will be made publicly available upon publication of the paper with a license that allows free usage for research purposes. (yes)
\item All source code implementing new methods have comments detailing the implementation, with references to the paper where each step comes from (yes)
\item If an algorithm depends on randomness, then the method used for setting seeds is described in a way sufficient to allow replication of results. (yes)
\item This paper specifies the computing infrastructure used for running experiments (hardware and software), including GPU/CPU models; amount of memory; operating system; names and versions of relevant software libraries and frameworks. (yes)
\item This paper formally describes evaluation metrics used and explains the motivation for choosing these metrics. (yes)
\item This paper states the number of algorithm runs used to compute each reported result. (yes)
\item Analysis of experiments goes beyond single-dimensional summaries of performance (e.g., average; median) to include measures of variation, confidence, or other distributional information. (no)
\item The significance of any improvement or decrease in performance is judged using appropriate statistical tests (e.g., Wilcoxon signed-rank). (no)
\item This paper lists all final (hyper-)parameters used for each model/algorithm in the paper’s experiments. (yes)
\item This paper states the number and range of values tried per (hyper-) parameter during development of the paper, along with the criterion used for selecting the final parameter setting. (NA)
\end{itemize}

\end{document}